\pdfoutput=1
\documentclass{article}

\usepackage{aaai21}  
\usepackage{times}  
\usepackage{helvet} 
\usepackage{courier}  
\usepackage[hyphens]{url}  
\usepackage{graphicx} 
\usepackage{natbib}  
\usepackage{caption} 
\usepackage{url}            
\usepackage{booktabs}       
\usepackage{amsfonts}       
\usepackage{nicefrac}       
\usepackage{microtype}      
\usepackage{mathtools}
\usepackage[vlined,boxed,commentsnumbered,linesnumbered,ruled]{algorithm2e}
\usepackage{subfigure}
\usepackage{dsfont}
\usepackage{amsmath}
\usepackage{amsthm}
\usepackage{graphicx}
\usepackage[switch]{lineno}
\DeclareMathOperator*{\argmax}{arg\,max}

\usepackage{amsmath}
\usepackage{amsfonts}
\usepackage{bm}

\usepackage{amsmath}
\newcommand{\tb}{\textbf}

\newcommand{\bx}{\mathbf{x}}
\newcommand{\bt}{\mathbf{t}}
\newcommand{\bX}{\mathbf{X}}
\newcommand{\bz}{\mathbf{z}}

\newcommand{\be}{\mathbf{e}}
\newcommand{\bth}[1][]{\theta_{#1}}
\newcommand{\bphi}{\phi}

\newcommand{\beps}{\bm{\epsilon}}
\newcommand{\bmu}{\bm{\mu}}

\newcommand{\bh}{\mathbf{h}}

\newcommand{\by}{\mathbf{y}}

\newcommand{\bZ}{\mathbf{Z}}

\newcommand{\pT}[1][]{p_{\bth[#1]}}
\newcommand{\qPhi}{q_{\bphi}}

\newcommand{\bbZ}{\mathbb{Z}}
\newcommand{\bbR}{\mathbb{R}}
\newcommand{\bbE}{\mathbb{E}}
\newcommand{\mcN}{\mathcal{N}}
\newcommand{\mcL}{\mathcal{L}}
\newcommand{\mcX}{\mathcal{X}}

\newcommand{\mcF}{\mathcal{F}}
\newcommand{\mcH}{\mathcal{H}}
\newcommand{\mcP}{\mathcal{P}}

\title{Synergetic Learning of Heterogeneous Temporal Sequences\\ for Multi-Horizon Probabilistic Forecasting}

\author {
    Longyuan Li\textsuperscript{\rm 1,2}\thanks{The first two authors have made equal contribution.},
    Jihai Zhang\textsuperscript{\rm 3}\footnotemark[1], 
    Junchi Yan\textsuperscript{\rm 1,3}\thanks{Corresponding authors are Junchi Yan and Yaohui Jin.}, 
    Yaohui Jin\textsuperscript{\rm 1,2 $\dagger$},\\
    Yunhao Zhang\textsuperscript{\rm 3},
    Yanjie Duan\textsuperscript{\rm 4}, 
    Guangjian Tian\textsuperscript{\rm 4}\\
}
\affiliations {
    \textsuperscript{\rm 1} MoE Key Lab of Artificial Intelligence, AI Institute, Shanghai Jiao Tong University \\
    \textsuperscript{\rm 2} State Key Lab of Advanced Optical Communication System and Network, Shanghai Jiao Tong University \\
    \textsuperscript{\rm 3} Department of Computer Science and Engineering, Shanghai Jiao Tong University\\
    \textsuperscript{\rm 4} Huawei Noah’s Ark Lab\\
    \{jeffli,yunfan243332345,yanjunchi,jinyh, zhangyunhao\}@sjtu.edu.cn,  \{duanyanjie1,tian.guangjian\}@huawei.com
}
\begin{document}
\maketitle

\begin{abstract}
    Time-series is ubiquitous across applications, such as transportation, finance and healthcare. Time-series is often influenced by external factors, especially in the form of asynchronous events, making forecasting difficult. However, existing models are mainly designated for either synchronous time-series or asynchronous event sequence, and can hardly provide a synthetic way to capture the relation between them. We propose Variational Synergetic Multi-Horizon Network (VSMHN), a novel deep conditional generative model. To learn complex correlations across heterogeneous sequences, a tailored encoder is devised to combine the advances in deep point processes models and variational recurrent neural networks. In addition, an aligned time coding and an auxiliary transition scheme are carefully devised for batched training on unaligned sequences. Our model can be trained effectively using stochastic variational inference and generates probabilistic predictions with Monte-Carlo simulation. Furthermore, our model produces accurate, sharp and more realistic probabilistic forecasts. We also show that modeling asynchronous event sequences is crucial for multi-horizon time-series forecasting.
\end{abstract}

\section{Introduction}
Temporal data streams are ubiquitous in areas including transportation, electronic health, economics, environment monitoring. One major type of temporal data is time-series (also known as synchronous sequence, e.g. temperature, economic index, ECG records), which consists of successive evenly-sampled discrete-time data points. Another type of temporal data is event sequence (also known as asynchronous sequence, e.g. e-commerce transactions, social interactions, extreme weathers), which consists of data points irregularly dispersed in the continuous-time domain~\cite{xiao2017modeling}. Both types carry rich information about the evolution of complex systems, based on which effective predictions are crucial to subsequent decision making, especially for those time-sensitive cases. Multi-horizon time-series forecasting has wide application scenarios that can be integrated into many business processes. One particular setting is to forecast long term multi-step future time-series as a whole. The hope is to avoid error accumulation that is common in single-step autoregressive models~\cite{benidis2020neural}.

While there have been recent works on modeling time-series~\cite{chandola2012anomaly,makridakis2018m4,mariet2019foundations,li2019enhancing,rangapuram2018deep,salinas2019deepar} and event sequence~\cite{du2016recurrent,bacry2015hawkes,xiao2016modeling,xiao2017modeling,wu2018decoupled}. Most of them are focused on one of these two types and the research in these two directions are conducted separately and independently. However, the two types of temporal data are usually strongly correlated and are likely to have lead-lag relationships. In other words, time-series is influenced by temporal events, and temporal events may also be caused or influenced by certain fluctuations in time-series. For example, traffic accidents may lead to high road occupancy rate, and constant high road occupancy rate over time can cause traffic accidents or traffic jams.

Since the two types of data are interrelated, using only one can possibly lead to deviation in prediction. We believe there is a real need to jointly model both types of data. There are related efforts in linking the synchronous time-series and asynchronous event sequences to each other. One way is to aggregate event sequences to fixed-interval count sequences to extract aligned time-series data. The other is to convert time-series to event sequences by extracting events from time-series based on human-defined rules~\cite{bacry2015hawkes}. However, such coarse treatments can lead to loss of key information about the actual dynamics of either two processes~\cite{binkowski2017autoregressive}.


This paper addresses the problem of jointly modeling time-series and event sequences for multi-horizon time-series forecasting. We propose Variational Synergetic Multi-Horizon Network (VSMHN). 
The basic idea is to learn the probability distribution of future values conditioned on heterogeneous past sequences (i.e. time-series and event sequence). 
Our model is based on conditional variational autoencoder (CVAE)~\cite{sohn2015learning}. 
We devise a hybrid recognition model which combines advances in deep point processes models and variational recurrent neural networks. The experimental results show that our model is able to untwine factors of asynchronous event sequence from time-series, and provides accurate and sharp multi-horizon probabilistic forecasting fulfilled by Monte-Carlo sampling. 

In summary, the main highlights of the paper are:

1) We argue the importance of capturing the interaction between event sequence and time-series in prediction. To our best knowledge, this is the first work that considers the problem of multi-horizon time-series forecasting with synchronous and asynchronous past temporal sequences.

3) Our model combines neural networks and probabilistic models. It can be trained by stochastic gradient descent and scales to large datasets. As the embodiment of both recognition model and prior model, the Time-Aware Hybrid RNN is carefully devised to preserve valuable timing information. 

3) We conduct experiments on public real-world datasets, showing its superiority against state-of-the-arts. We find event sequence is crucial to accurate multi-horizon forecasting, even when events themselves are indirectly extracted from the time-series. This also indicates the utility of our approach for multi-dimensional time-series data.

\section{Related Works}
\paragraph{Variational Autoencoder (VAE).}
VAE~\cite{kingma2013auto} is an efficient way to handle latent variables in neural networks. The VAE learns a generative model of the marginal probability $\pT(\bx)$ of the observed data $\bx$. The generative process of the VAE is as follows: A set of latent variable $\bz$ is generated from the prior distribution $\pT(\bz)$ and the data $\bx$ is generated by the generative model $\pT(\bx|\bz)$ conditioned on $\bz$. A recognition model $\qPhi(\bz|\bx)$ is introduced to approximate the true posterior $\pT(\bz|\bx)$. Further, the conditional variational autoencoder (CVAE)~\cite{sohn2015learning} extends the VAE to model conditional generative process $\pT(\by|\bx)$ of target $\by$ given features $\bx$.

\paragraph{Time-series forecasting.}
Traditional statistical forecasting models include linear autoregressive models, such as ARIMA~\cite{box2015time}, Exponential Smoothing~\cite{hyndman2008forecasting} and VAR~\cite{lutkepohl2005new}. They are well-understood and still competitive in many forecasting competitions~\cite{makridakis2018m4}. Deep neural networks have been proposed to learn from multiple related time-series by fusing traditional models, such as DeepAR~\cite{salinas2019deepar}, RNN and Gaussian copula process model~\cite{salinas2019high}, Deep State Space models~\cite{rangapuram2018deep} and its interpretable version \cite{li2019learning}, Deep Factor models \cite{wang2019deep}. 

Another line of network architectures for multi-horizon forecasting involves sequence-to-sequence models~\cite{sutskever2014sequence,fan2019multi}, which are powerful tools in the domain of Natural Language Processing (NLP), and are often used in machine translation tasks. Rather than modeling each time point within a sequence in an autoregressive manner, the sequence-to-sequence models use an 'encoder-deocoder' framework, to learn a mapping between arbitrarily long sequences through an intermediate encoded state. Its loss~\cite{vincent2019shape}, evaluation metrics~\cite{taieb2015bias} and generalization bounds~\cite{mariet2019foundations} are well-studied in the context of multi-horizon time-series forecasting.

The methods mentioned above are for evenly-sampled synchronous time-series. Models for other types of time-series such as multi-rate time-series~\cite{che2018hierarchical} and irrgularly-sampled time-series~\cite{binkowski2017autoregressive,baytas2017patient} are developed, however, they cannot generalize to model heterogeneous temporal data.

\paragraph{Temporal Point Processes (TPPs).}
Different from evenly-sampled time-series, the timestamps of events carry rich information of  dynamics. TPPs are powerful tools for modeling such continuous-time event sequence~\cite{wu2018decoupled}. Recently, neural point processes such as Recurrent Point Process~\cite{xiao2017modeling}, Recurrent Marked Temporal Point process~\cite{du2016recurrent} and Neural Hawkes process~\cite{mei2017neural} are broadly discussed. \cite{xiao2019learning} jointly learns event sequences with time-series. However, like other models mentioned above, it can only predict the timestamp and type of the next event, and thus is not suitable for our multi-horizon time series forecasting problem.

One alternative is to convert event sequences to dummy-coded time-series aligned with others (0/1 sequence for each event type). However, this approach has obvious drawbacks: 1) The converted event sequence is sparse and contains many zeros between events; 2) The dimension equals to the number of event types, making the sequence even sparser; 3) The information of the intervals between events is lost, especially when more than one event occurs in a single interval.

Based on the observations above, if we model such data using neural networks such as GRUs or LSTMs, the required non-linearity of the model and the amount of information are imbalanced: The model need to have many layers to memory the previous event after many superfluous transitions, while an asynchronous approach just need to memory last event and type as they are. In conclusion, a hybrid model is welcomed to fully utilize information and learn correlations from heterogeneous temporal sequences.
\begin{figure}[tb!]
\centering
\includegraphics[width=0.48\textwidth]{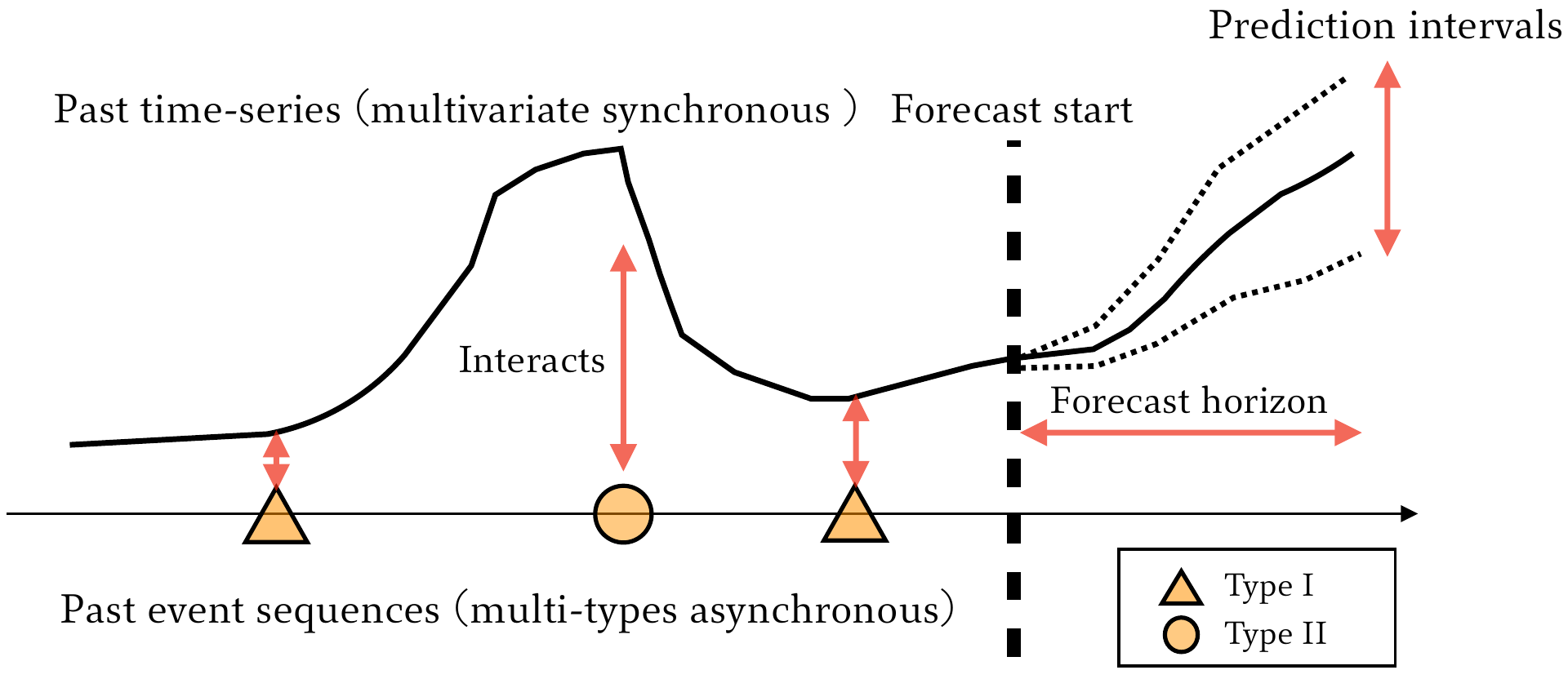}
\caption{Multi-horizon probabilistic forecasting for heterogeneous sequences. As past time-series is influenced by events, a joint model is used to untwine the exogenous factors. Event sequence is unknown in forecasting horizon, and we are interested in confidence intervals of predictions.}
\label{fig:problem}
\end{figure}
\section{Variational Synergetic Multi-Horizon Network}
\paragraph{Problem Formulation.}\label{sec:problem} 
Denote an $M$-dimensional multivariate evenly-sampled time-series having $T$ time steps: $\mcX_{1:T} = \{ \bx_1, \bx_2,\dots,\bx_T\}$ where each $\bx_t\in \bbR^M$. A corresponding irregularly-sampled event sequence can be represented as $\{(c_1,t_1),(c_2,t_2),\dots (c_n,t_n)\}$, where $c_i \in \{1,\dots,C\}$ denotes the class of the $i$-th event and $t_i$ is the time of occurrence, and $t_n\leq T$. We assume the events arrive over time, i.e. $\{t_i\in \bbR | t_i>t_{i-1}\}$. Let $\mcH_t=\{(c_i,t_i)|t_i< t\}$ denote the event history until $t$. Our goal is to estimate the distribution of time-series in the future window $\tau\in\mathbb{N}^+$ given history of time-series and event (see Fig.~\ref{fig:problem}):
\begin{equation}\label{eq:problem}
    p(\mcX_{T+1:T+\tau}|\mcX_{1:T}, \mcH_{T},\Phi)
\end{equation}
where $\Phi$ denotes the model parameters. Note that time-series starts from $t=1$, and event sequence start from $t=0$. Note that different from exogenous variables of other models, event sequence is unknown in the forecast horizon.

\paragraph{Objective.} 
The  forecasting quality is generally measured by a metric between the predicted and actual values in the forecast horizon. Probabilistic forecasts estimate a probability distribution rather than predicting a single value as point forecasts do. Simple accuracy metrics such as MAE, RMSE are considered incomplete because they are unable to evaluate uncertainties. The {Continuous Ranked Probability Score} (CRPS) generalizes the MAE to evaluate probabilistic forecasts~\cite{gneiting2014probabilistic}. Given the true observation $x$ and the cumulative distribution function (CDF) of its forecasts distribution $F_X$, the CRPS is given by:
\begin{equation}
    \text{CRPS}(F_X, x) = \int_{-\infty}^{\infty}\left(F_X(y)-\mathds{1}(y-x)\right)^2\mathrm{d}y
\end{equation}
where $\mathds{1}$ is the Heaviside step function.

\begin{figure*}[tb!]
\centering
\includegraphics[width=0.86\textwidth]{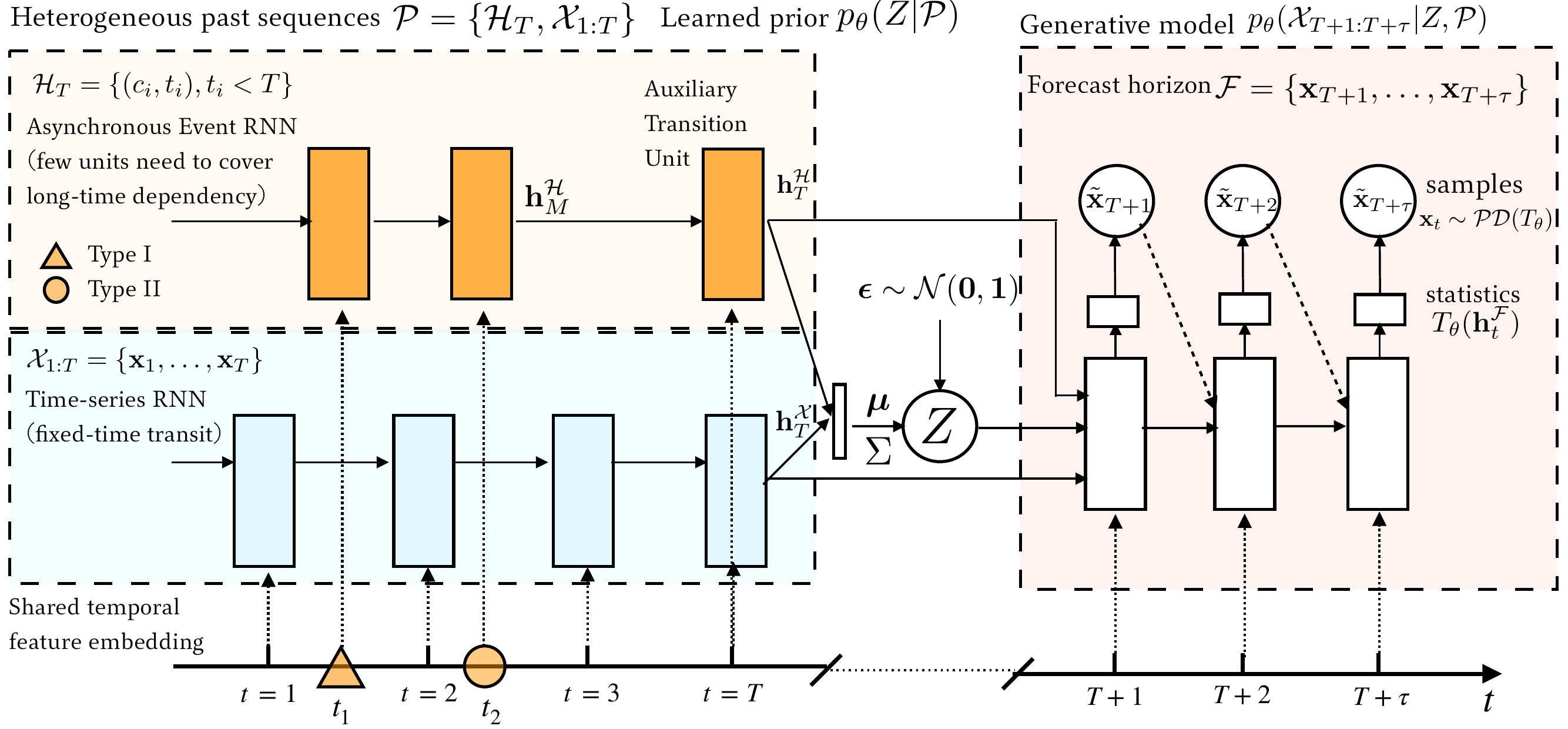}
\caption{Overview of our synergetic model for probabilistic multi-horizon time series forecasting. In the plot, we set history length $T=4$, and $M=2$ event occurrences within the time range, the forecast horizon $\tau=3$. The left side is the prior model $\pT(\bZ|\mcP)$. In training, the recognition model (not shown in the plot) shares the same parameters of the time-series RNN with the prior model. The recognition model transits $\tau$ more times with ground truth future value input and then outputs $\bh_{T+\tau}^{\mcX}$. It aims to maximize the conditional probability $p(\mcF|\mcP)$ of future value $\mcF$ given past $\mcP$, by minimizing the sum of KL-divergence between $\qPhi(\bZ|\mcP,\mcF)$ and $\pT(\bZ|\mcP)$ and negative log-likelihood of future data $\pT(\mcX_{T+1:T+\tau}|\bZ, \mcP)$.}
\label{fig:arch}
\end{figure*}

\paragraph{Model Architecture.}
Our task is to model the probability distribution of future time-series $\mcF=\mcX_{T+1:T+\tau}$ given heterogeneous past data $\mcP=\{\mcX_{1:T}, \mcH_T\}$. We denote the modeled conditional distribution by $p(\mcF|\mcP)$. We introduce a latent variable $\bZ$ conditioned on the observed data, and a variational autoencoder conditioned on $\mcP$ (CVAE~\cite{sohn2015learning}) is trained to maximize the conditional log likelihood of $\mcF$ given $\mcP$, which involves an intractable marginalization over the latent variable $\bZ$, i.e.:
\begin{equation}
    p(\mcF|\mcP) = \int_{Z}p(\mcF,\bZ|\mcP)\,\mathrm{d}\bZ = \int_{Z}p(\mcF|\mcP,\bZ)p(\bZ|\mcP)\,\mathrm{d}\bZ
\end{equation}

The CVAE can be trained by maximizing the {variational evidence lower bound (ELBO)} of the conditional log likelihood $\mcL(\theta,\phi)$, which is derived as:
\begin{equation}\label{eq:elbo}
\begin{split}
    \log \pT(\mcF|\mcP)&= \text{KL}\left(\qPhi(\bZ|\mcP, \mcF)\rVert\pT(\bZ|\mcP,\mcF)\right)\\
    &+\bbE_{\qPhi}\left[-\log \qPhi(\bZ|\mcP,\mcF)+\log\pT(\mcF,\bZ|\mcP) \right]\\
    &\ge -\text{KL}(\qPhi(\bZ|\mcP, \mcF)\rVert\pT(\bZ|\mcP)) \\
    &+ \bbE_{\qPhi}\left[\log \pT(\mcF|\mcP,\bZ) \right]
    \coloneqq \mcL(\theta, \phi)
\end{split}
\end{equation}
where $\qPhi(\bZ|\mcP, \mcF)$ is simplified by notation $\qPhi$, and $KL$ denotes Kullback-Leibler (KL) divergence. A proposal distribution $\qPhi(\bZ|\mcP, \mcF)$ parameterized by $\phi$, which is known as a `recognition' model or probabilistic `encoder', is introduced to approximate the true posterior distribution $\pT(\bZ|\mcP, \mcF)$. The future time-series distribution is generated from the distribution $\pT(\mcF|\mcP,\bZ)$, which is also known as `generative' model or probabilistic `decoder'. The prior distribution $\pT(\bZ|\mcP)$ of latent variable $\bZ$ is modulated by the past $\mcP$.

Then we need to parameterize the three distributions and optimize $\mcL$ with respect to $\theta$ and $\phi$. The KL term of Eq.~\ref{eq:elbo} can be analytically computed assuming a simple distribution of $\bZ$, typically Gaussian, while the second expectation term cannot, making $\mcL$ not differentiable. To obtain a differentiable estimation of ELBO, we resort to Stochastic Gradient Variational Bayes (SGVB)~\cite{kingma2013auto}. The basic idea is to sample from an auxiliary stochastic variable instead directly from the recognition model $\qPhi$. As such, the gradient of the objective can be back-propagated through the sampled $\bZ$. Formally we have the empirical lower bound:
\begin{equation}\label{eq:SGVB_ELBO}
\begin{split}
    \tilde{\mcL}_{\text{SGVB}}(\mcP,\mcF;\theta,\phi)
    &= -\text{KL}\left(\qPhi(\bZ|\mcP, \mcF)\rVert\pT(\bZ|\mcP)\right) \\
    &+ \frac{1}{K}\sum_{k=1}^{K}\log \pT\left(\mcF|\mcP,\bZ^{(k)}\right)
\end{split}
\end{equation}
where $\bZ^{(k)}=q_{\phi}(\mcP,\mcF,\beps^{(k)})$, $\beps^{(k)}\sim\mcN(\mathbf{0},\mathbf{I})$ and $K$ is the number of samples.

\paragraph{Remarks.}
Compared with a direct deterministic mapping such as $\mcF=f_{\theta}(\mcP)$, the stochastic latent variable $\bZ$ allow for modeling multiple modes in conditional distribution of future time-series $\mcF$ given past data $\mcP$, making the proposed model able to model one-to-many mapping, which is common in time-series forecasting tasks e.g. the same past data lead to different future data due to concept drift~\cite{gama2014survey}.

Moreover, in other probabilistic mapping approaches, such as quantile regression loss MQRNN~\cite{wen2017multi} or likelihood models~\cite{salinas2019deepar}, the randomness only exists in the expectation term $\bbE$ in Eq.~\ref{eq:elbo}, and thus can be viewed as maximum likelihood approaches. In contrast, our model marginalizes the latent variable, and the KL-term can be viewed as a regularization term. As a result, compared with maximum likelihood approaches, our approach has the potential to learn better uncertainty of the data distribution and be more robust on small datasets.

\paragraph{Implementations.}
As previously discussed, the time-series is evenly-sampled discrete-time synchronous data, while the alongside event sequence is continuous-time asynchronous data, making it difficult to model them jointly, especially to preserve the actual time order, which is essential for causality, i.e. past data cannot be influenced by future data. To jointly learn representations of heterogeneous sequences, we introduce the Time-aware Hybrid RNN as a building block of recognition model and prior model as shown in Fig.~\ref{fig:arch}

\textbf{Time-Aware Hybrid RNN.} Suppose an input of past time-series $\bX_{1:T}$ with strict $T$ observation vectors and event sequences $\mcH_T$, where there are $M\in\bbZ^{\ge0}$ events within the past range $(0,T]$. Firstly, we use an RNN $f_{\phi}$ to model the time-series, which is transited at fixed time \{1,\dots,T\}. For asynchronous events, motivated by \cite{du2016recurrent}, we build an RNN $g_{\phi}$ that transits at each event, and takes timing and embedded event type as input. Moreover, to learn them in the same timeline, we add a shared feature extractor of time-relevant features to both RNNs. Typical features include absolute time, hour of days, etc. Specifically:
\begin{equation}\label{eq:TARNN}
\begin{split}
    \bh^{\mcX}_t &= f_{\phi}\left([\varphi(\bx_t), \psi(\bt_t)], \bh_{t-1}^{\mcX}\right)\quad\quad\quad\quad\ \text{for}\; t=1\dots T  \\
    \bh^{\mcH}_m &= g_{\phi}\left([\zeta(\be_m), \bigtriangleup t_m, \psi(\bt_m)],\bh_{m-1}^{\mcH}\right)\quad \text{for}\;m=1\dots M
\end{split}
\end{equation}
where $[a, b]$ denotes the concatenation of $a$ and $b$, $\bigtriangleup t_m=t_m-t_{m-1}$ denotes the inter-event duration, and $\be$ denotes one-hot coded event type vector. The bold $\bt_t$ denotes the vector of temporal features, and $\psi$ is the shared feature extractor, parameterized by MLP. $\varphi$ and $\zeta$ are feature extractors of time-series and events, also parameterized by MLP. Both RNNs are initialized with zero vector $\bh$.

\textbf{Auxiliary Transition Unit.} When modeling very long sequences, a typical practice is to split time-series into chunks that are overlapped at the time axis \cite{salinas2019deepar}. However, with the introduction of event sequences, such an approach causes a problem. Because events occur irregularly, many adjacent chunks may share the same set of events, where the event type, timing, and all other features are the same. However, the forecast targets can be different for the adjacent chunks, which means that the same event sequence input may reflect different targets, making it difficult for the model to capture features within the event sequence. To solve the problem, we introduce Auxiliary Transition Unit, whose basic idea is to let the asynchronous RNN know the exact end time as the time-series RNN, which is done by an auxiliary transit at $T$ with zero event type input:
\begin{equation}
    \bh_{T}^{\mcH}=g_{\phi}\left([\zeta(\mathbf{0}), \bigtriangleup t_T, \psi(\bt_T)],\bh_{M}^{\mcH}\right)
\end{equation}
where $\mathbf{0}$ denotes zero vector which has the same size as $\be$, and $\bigtriangleup t_T=t_T-t_M$. By doing so, our model can be trained in batch effortlessly, without worrying about data conflicts.

\textbf{Synergetic Layer.} At time $T$, we have extracted features $\bh^{\mcX}_{T}$ and $\bh^{\mcH}_T$ by two RNNs from heterogeneous sequences. Then the problem is to parameterize the recognition model $\qPhi(\bZ|\mcP, \mcF)$ and prior model $\pT(\bZ|\mcP)$. We assume the latent variable follows diagonal Gaussian distribution $\bZ\sim \mcN(\bmu, \Sigma)$. For the prior model $\pT(\bZ|\mcP)$, we take the following parameterization:
\begin{equation}\label{eq:prior}
\begin{split}
\bmu &= \text{MLP}_{\theta}\left(\bh^{\mcX}_{T}, \bh^{\mcH}_T\right),\\
\log(\Sigma) &= \text{diag}\left(\text{MLP}_{\theta}(\bh^{\mcX}_{T}, \bh^{\mcH}_T)\right)
\end{split}
\end{equation}
where the inputs of MLPs are concatenated vector. For the recognition model $\qPhi(\bZ|\mcP, \mcF)$ that depends on both past data $\mcP$ and future target $\mcF$, we just need to let the same time-series RNN transit $\tau$ more times, and the extracted feature is denoted by $\bh_{T+\tau}^{\mcX}$. The parameterization of recognition model $\qPhi(\bZ|\mcP, \mcF)$ is:
\begin{equation}\label{eq:prior2}
\begin{split}
\bmu =& \text{MLP}_{\phi}\left(\bh^{\mcX}_{T+\tau}, \bh^{\mcH}_T\right),\\
\log(\Sigma) =& \text{diag}\left(\text{MLP}_{\phi}(\bh^{\mcX}_{T+\tau}, \bh^{\mcH}_T)\right)
\end{split}
\end{equation}

\textbf{Stochastic RNN Generation Model.} Given sampled $\bZ\sim \qPhi(\bZ|\mcP,\mcF)$ and inferred $\bh^{\mcX}_{T}$ and $\bh^{\mcH}_T$, the task becomes how to parameterize $\pT(\mcF|\mcP,\bZ)$. We do not assume any particular distribution of the generated data $\bx$, but instead we assume it follows a parametric distribution $\mathcal{PD}$ such as ones in the exponential family, and it has sufficient statistic $\mathbf{T}$. Specifically, we initialize an RNN $d_{\theta}$ with:
\begin{equation}\label{eq:p_init}
    \bh_T^{\mcF} = \text{MLP}_{\theta}^{1}\left([\bh^{\mcX}_{T}, \bh^{\mcH}_T, \text{MLP}^{2}_{\theta}(Z)]\right)
\end{equation}
where the inner $\text{MLP}^2$ is a simple feed-forward network to map sampled $\bZ$ as a concatenation with hidden state $\bh$. The outside $\text{MLP}^1$ is to map the concatenated feature to the hidden dimension of the decoder RNN. Then, we iteratively compute from $T+1$ to $T+\tau$:
\begin{equation}
\begin{split}
    \bh_t^{F} = d_{\theta}\left([\varphi(\bx_{t-1}), \psi(\bt_t)], \bh_{t-1}^{F}\right),
    \quad
    \bx_{t}\sim \mathcal{PD}\left(\mathbf{T}_{\theta}(\bh_{t}^{F})\right)
\end{split}
\label{eq:observation}
\end{equation}
where $\varphi$ and $\psi$ are the shared feature extractor of observation and temporal features respectively. Note that $\bx_T$ is true value from past data $\mcX$. 

For $\mathbf{T}_{\theta}(\bh_t^{\mcF})$, we also use networks to parameterize the mapping. In order to constrain real-valued output $\bh_t^{\mcF}$ to the parameter domain, we use the following transformations:
\begin{itemize}
    \item Real-valued parameters: no transformation.
    \item Positive parameters: the $\text{SoftPlus}$ function.
    \item Bounded parameters $[a,b]$: scale and shifted $\text{Sigmoid}$ function $y = (b-a)\frac{1}{1+\exp(-\tilde{y})}+a$
\end{itemize}

So far, the recognition model $\qPhi(\bZ|\mcP, \mcF)$, prior model $\pT(\bZ|\mcP)$, and generation model $\pT(\mcF|\mcP,\bZ)$ are fully specified. They can be trained by Eq.~\ref{eq:SGVB_ELBO} w.r.t $\theta$ and $\phi$.

\subsection{Forecasting with Confidence Intervals}
\begin{algorithm}[tb!]
\caption{Forecasting by Monte-Carlo sampling}\label{alg:forecast}
\KwIn{Heterogeneous past data $\mcP=\{\mcX_{1:T}, \mcH_T\}$;}
\KwIn{Trained model $\pT(\mcF|\mcP,\bZ)$ and $\pT(\bZ|\mcP)$;}
\KwIn{Forecast horizon $\tau$ and number of samples $N$;}
Evaluate Eq.~\ref{eq:TARNN} to get $\bh^{\mcX}_{T}$ and $\bh^{\mcH}_T$;\\
Compute $\pT(\bZ|\mcP)=\mcN(\bmu, \Sigma)$ by Eq.~\ref{eq:prior};\\
\For{$n=1$ to $n=N$}{
    Sample $\bZ\sim \mcN(\mu,\Sigma)$;\\
    $\bh_T^{\mcF} = \left[\bh^{\mcX}_{T}, \bh^{\mcH}_T, \text{MLP}(Z)\right]$;\\
    \For{$t=T+1$ to $T+\tau$}{
    $\bh_t^{F} = d_{\theta}\left([\varphi(\bx_{t-1}), \psi(\bt_t)], \bh_{t-1}^{F}\right)$;\\
    $\bx_{t}^{(n)}\sim \mathcal{PD}\left(\mathbf{T}_{\theta}(\bh_{t}^{F})\right)$;
    }
}
Compute mean and quantiles from sampled $\bx_t^{(n)}$, $n=1\dots N$ for each time $t$;
\end{algorithm}
Once the model is learned, we can use the model to address the forecasting problem addressed in Eq.~\ref{eq:problem} for new coming time-series and event sequences. The straightforward way is to perform a deterministic inference without sampling $\bZ$, i.e., $\mcF^{*}=\argmax_{\mcF}\pT(\mcF|\mcP,\bZ^{*}),\ Z^{*}=\bbE[\bZ|\mcP]$. However, since our transition model is an RNN, which is non-linear, computing of multi-horizon future values is difficult. We would like to make probabilistic forecast, so we resort to a Monte-Carlo sampling approach as Algorithm~\ref{alg:forecast}.

\section{Experiments}
\begin{table*}[tb!]
\centering
    \resizebox{0.9\linewidth}{!}{
\begin{tabular}{lcrrrrrr}
\hline
Method     & Event used?   & PM2.5   & Dewpoint    & Temperature     & Pressure   & Electricity    & Traffic     \\ \hline
DeepAR    && 85.93/45.57     & 6.27/3.48     & 2.65/1.47     & 50.16/39.92        & 23.9/11.34     & 1.52/0.71     \\
DeepAR-E     & \checkmark & 88.82/48.44& 5.86/3.21     & 3.10/1.79     & 35.95/28.00        & 25.23/12.50    & 1.55/0.70     \\
MQRNN&& 116.94/70.48    & 8.72/6.86     & 3.46/2.35     & 6.88/4.81     & 27.00/14.01    & 1.50/0.74     \\
DeepFactor && 145.67/112.58        & 10.02/7.80    & 4.98/3.97     & 6.33/5.05     & 29.38/15.56    & 2.32/1.59     \\
\hline
VAR & &88.39/46.18 & \tb{3.57}/2.01 & 3.09/1.70 &2.91/1.93 &21.62/11.92 &1.73/1.43  \\
ETS & &107.06/55.34 & 4.26/2.19 &3.15/1.81 &3.19/1.87 &21.14/11.51 &2.62/2.95  \\
SARIMA & &92.44/47.57 & 3.98/2.06 &2.85/1.68 &4.58/2.50 &22.69/12.05 &1.70/0.86  \\ \hline
VSMHN-TS (ours)      &            & 78.47/36.98     & 4.12/2.05     & 2.16/1.20     & 2.91/1.53     & 20.51/9.21     & 1.68/0.53     \\
VSMHN (ours)      & \checkmark & \tb{72.84}/\tb{33.67} & 3.59/\tb{1.94} & \tb{2.10}/\tb{1.19} & \tb{2.47}/\tb{1.33} & \tb{19.08}/\tb{8.86} & \tb{1.36}/\tb{0.48} \\ \hline
\end{tabular}}
\caption{RMSE/CRPS comparison of rolling-day forecast of last 28 days. Tick denotes event information is used in the model.}
\label{tab:performance}
\end{table*}
To provide evidence for the practical effectiveness of our proposed VSMHN model, we conduct experiments with real-world datasets. We also compare to a wide range of machine learning models for multi-horizon forecasting. 

We implement our model by Pytorch on a single RTX-2080Ti GPU. We set the dimension of hidden layers of all MLPs and RNNs in the generative model and inference model to 100, and the dimension of stochastic latent variable $\bZ$ to 50. We set sample size $K$ of the objective Eq. \ref{eq:SGVB_ELBO} to 5. We use Adam optimizer with learning rate 0.001. Forecast distribution is estimated by 1000 trails of Monte Carlo sampling. We use Gaussian observation model (Eq.~\ref{eq:observation}) where mean and standard deviation are parameterized by two layer NNs. Additional details about hyper-parameter optimization are given in the supplementary materials.
\subsection{Datasets and Protocols.}
We use two univariate datasets and a multivariate dataset. For all datasets, we extract 4 temporal features: absolute time, hour-of-day, day-of-week, and month-of-year for hourly-sampled time-series data. We held-out the last four weeks for evaluation, and the rest for training. We train the model to predict the future day (24 data points) given the past week (168 points), along with events within that week.

\textbf{Electricity.} The UCI household electricity dataset contains time-series of the electricity usage in $kW$ recorded hourly for 370 clients. We work on an univariate series of length 21044, and time points with fluctuations larger than 30 are extracted as events, where the event types are up and down accroding to the direction.

\textbf{Traffic.} The dataset corresponds to hourly-sampled road occupancy rates in percentiles (0-100\%) from the California Department of Transportation, we work on the first univariate series of length 17544. We extract time points with fluctuations larger than 10\% as events, whose types are up and down according to the direction.

\textbf{Environment.} A public air quality multivarite dataset~\cite{li2019learning}. We are interested in how atmospheric variables interact with weather events. We use four hourly-sampled variables: PM2.5, dew point, temperature and pressure. And we extract three types of events from minute-level data sources: wind start, wind stop and rain start.
\paragraph{Data Preprocessing.}
Given full training time-series $\mcX=\{\bx_1,\dots,\bx_T\}$ and alongside event sequence $\mcH_T$, where $T$ is large, we need to generate multiple training samples $\{\mcP^{(i)},\mcF^{(i)}\}_i{i=1}^{N}$. For the time-series part, we follow the protocol used in~\cite{salinas2019deepar}: The \textit{shingling} technique is to convert long time-series into multiple chunks. Specifically, at each training iteration, we sample a batch of windows with width $W$ at random start point $[1,T-W+1]$ from the dataset and feed into the model. For the event sequence part, we first difference the timestamps to create inter-event duration feature $\bigtriangleup t$, and then join the event sequence to time-series chunks by the shared time index. When training, event sequences are padded ahead with zeros to the same length as a common practice in neural point process models~\cite{xiao2016modeling}. Moreover, we generate four time features (similar to positional encoding mechanism in Transformer~\cite{vaswani2017attention}): \textit{absolute-time}, \textit{hour-of-day}, \textit{day-of-week} and \textit{month-of-year} using the observation time stamps. To facilitate training, each of the variables in the time series is scaled to $\mcN(0,1)$ respectively to make their log likelihood comparable.

\paragraph{Compared Baselines.}
We compare three network models. 

\textit{DeepAR}~\cite{salinas2019deepar}, an RNN-based autoregressive likelihood model, which is essentially an univariate model. Multivariate time-series is trained jointly with static category features. It can learn correlations of target time-series with synchronous exogenous sequences, which are assumed to be known in the forecast horizon.

 \textit{DeepFactor}~\cite{wang2019deep}, a deep global-local model, in which the global model is an RNN, and the local model is a probabilistic time-series model such as Gaussian Processes or linear dynamical system.

 \textit{MQRNN}~\cite{wen2017multi}, a sequence-to-sequence model with LSTM encoder and decoder. The model is trained with quantile loss to learn uncertainties.

Neural network-based baseline are implemented by \textit{Gluon-ts} python package, which is based on MXNet, and hyperparameters are tuned using grid search. Moreover, to evaluate the influence of event input, we set DeepAR-E, which takes dummy-coded event sequences as exogenous variables that are unknown in the forecast horizon. We also use VSMHN-TS i.e. a pure time-series model that takes no events input. 

Recent study~\cite{makridakis2018m4} shows that statistical models are still competitive in time series forecasting, of which three are compared.

\textit{Vector Autoregression (VAR)}, a model that learns a linear mapping between the value-vectors and their lagged value-vectors in multi-variate time series data.

 \textit{Seasonal Autoregressive Integrated Moving Average (SARIMA)}, an additive model with autoregressive, differencing ,moving average and seasonal terms.

\textit{Exponential Smoothing (ETS)}, a model using exponential functions to weight past observations in evolution functions with exponentially decreasing weights over time. Particularly, we use Holt's Winters Seasonal Exponential Smoothing with an additive trend component, an additive seasonal component and an additive error component. 

The VAR and ETS models are implemented using \textit{statmodels} pyhton package. The SARIMA model is implemented using \textit{pmdarima} python package. The seasonal period hyper-parameter of ETS and SARIMA is set to 24 since the used data are hourly sampled. Other hyper-parameters are tuned by grid search according to Akaike Information Criterion (AIC) on training data.

\paragraph{Evaluation Metric}
CRPS is calculated analytically by the \textit{properscoring} python package. We calculate Root Mean Square Error (RMSE) using the mean of forecast distribution. All scores are calculated on the original scale of data.
\begin{figure}
    \centering
    \includegraphics[width=0.48\textwidth]{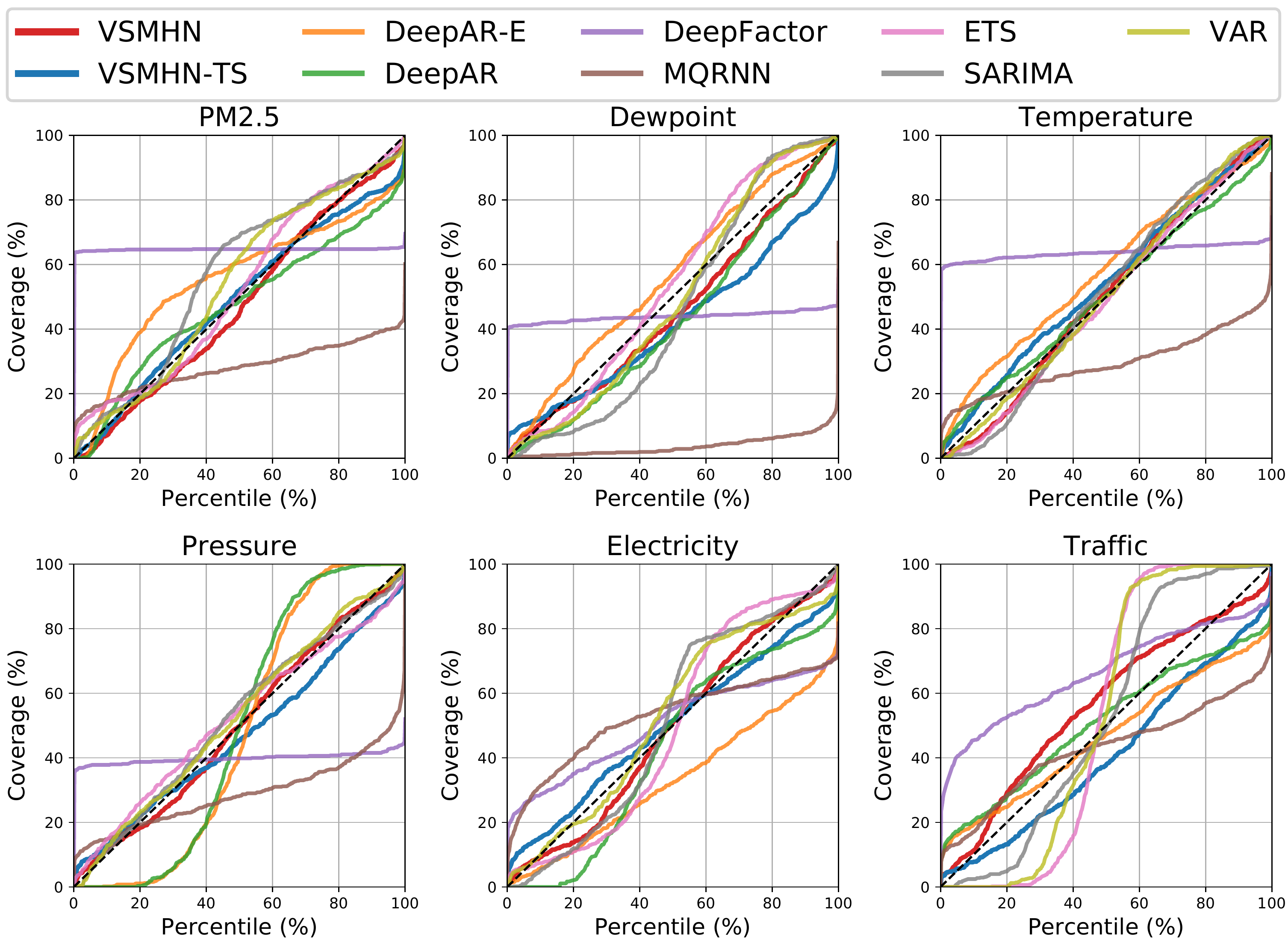}
    \caption{Uncertainty calibration. Perfect calibration corresponds to $\text{Coverage}(p)=p$ in diagonal black dotted line.}
    \label{fig:calibration_curve}
\end{figure}

We use Gaussian observation model, and CRPS can be computed exactly~\cite{gneiting2005calibrated}. Consider the predicted distribution is $\mcN(\mu,\sigma^2)$, the CRPS score is given by:
\begin{equation}
\begin{split}
&\text{crps}[\mcN(\mu,\sigma^2),x]=\\
&\sigma\Big \{ \frac{x-\mu}{\sigma}\big [2\Phi(\frac{x-\mu}{\sigma})-1\big ] +2\varphi(\frac{x-\mu}{\sigma})-\frac{1}{\sqrt{\pi}}\Big \}
\end{split}
\end{equation}
where $\Phi(\frac{x-\mu}{\sigma})$ and $\varphi(\frac{x-\mu}{\sigma})$ denote PDF and CDF, respectively, of the normal distribution with mean 0 and variance 1. For time series with $M$ variables,  the average is:
\begin{equation}
    \text{CRPS}=\frac{1}{M}\sum_{i}\text{crps}(F_{i},x_i)
\end{equation}

Note that the CRPS score reduces to MAE if each $F_{i}$ is a deterministic forecast.

The RMSE is defined as the rooted mean squared error over all time series, i.e., $i=1,\dots,M$, and over the forecast horizon, i.e. $t=T+1,T+2,\dots,T+\tau$:
\begin{equation}
    \text{RMSE} = \sqrt{\frac{1}{M\times\tau}\sum_{i,t}(x_{i,t}-\hat{x}_{u,t})^2}
\end{equation}
where $x$ is the true value in the forecast horizon, and $\hat{x}$ is the predicted distribution mean.
\subsection{Results and Discussion}
\paragraph{Overall accuracy.}
The RMSE and CRPS results are shown in Table~\ref{tab:performance}. Our proposed model (VSMHN with event input) beats state-of-the-art baselines on five of six variables from three datasets with respect to RMSE score. Our model outperforms all baselines on CRPS score notably, suggesting our model can provide sharp and accurate probabilistic forecast.

\textbf{Uncertainty calibration.}
For probabilistic forecast, we obtain not only values but also their probability distributions to assist decision-making. It is important to measure how well the forecast distribution is calibrated against the forecast error. We use coverage metric $\text{Coverage}(p)$, where $p$ is the percentile of the distribution. $\text{Coverage}(p)$ is the frequency of actual values lying within percentile $p$. Being closer to the perfect calibration line $\text{Coverage}(p)=p$ means better uncertainty estimation. Figure~\ref{fig:calibration_curve} shows the calibration curves of the forecast distribution. To quantify uncertainty calibration performance, we also compute the R-squared ($R^2$) values between the models' uncertainty calibration curves and the perfect calibration line, as shown in Table~\ref{tab:calibration_performance}. In conclusion, compared with various baselines on six variables, our model provides better uncertainty calibration.

\textbf{Impact of event sequence.}
To illustrate the influence of event sequence input to time-series, we convert dummy coded event sequence to time-series, and let DeepAR take it as exogenous input. We find event inputs can slightly improve the performance of DeepAR on some datasets. For our proposed VSMHN model, event sequence input constantly improves the performance, indicating that our model extracts valuable features from event sequence. Those features can help the model explain fluctuations within time-series in the training phase, thus improve overall forecast accuracy.

\textbf{Impact of model architecture.}
Surprisingly, we find that our model is also ahead of baselines in the absence of event inputs, which means it is still advanced even for conventional multi-horizon time-series forecasting problem as in~\cite{salinas2019deepar}. We find that it is difficult for DeepAR to forecast pressure over other variables as shown in Table~\ref{tab:performance}, which we think is because that as a univariate model, DeepAR cannot handle multivariate time-series with a wide range of scales properly. In contrast, our model is multivariate and performs stable on variables with different scales.

\textbf{Performance of traditional statistical models.} 
In Table~\ref{tab:performance}, traditional statistical models are competitive. However, simple statistical models struggle to extract features from complex external information or do cross-learning from multivariate data. As shown in~\cite{bojer2020kaggle}, successful integration of external information can be key to accurate forecast on complex time series datasets.

\begin{figure}[!tb]
    \centering
    {\includegraphics[width=0.48\textwidth]{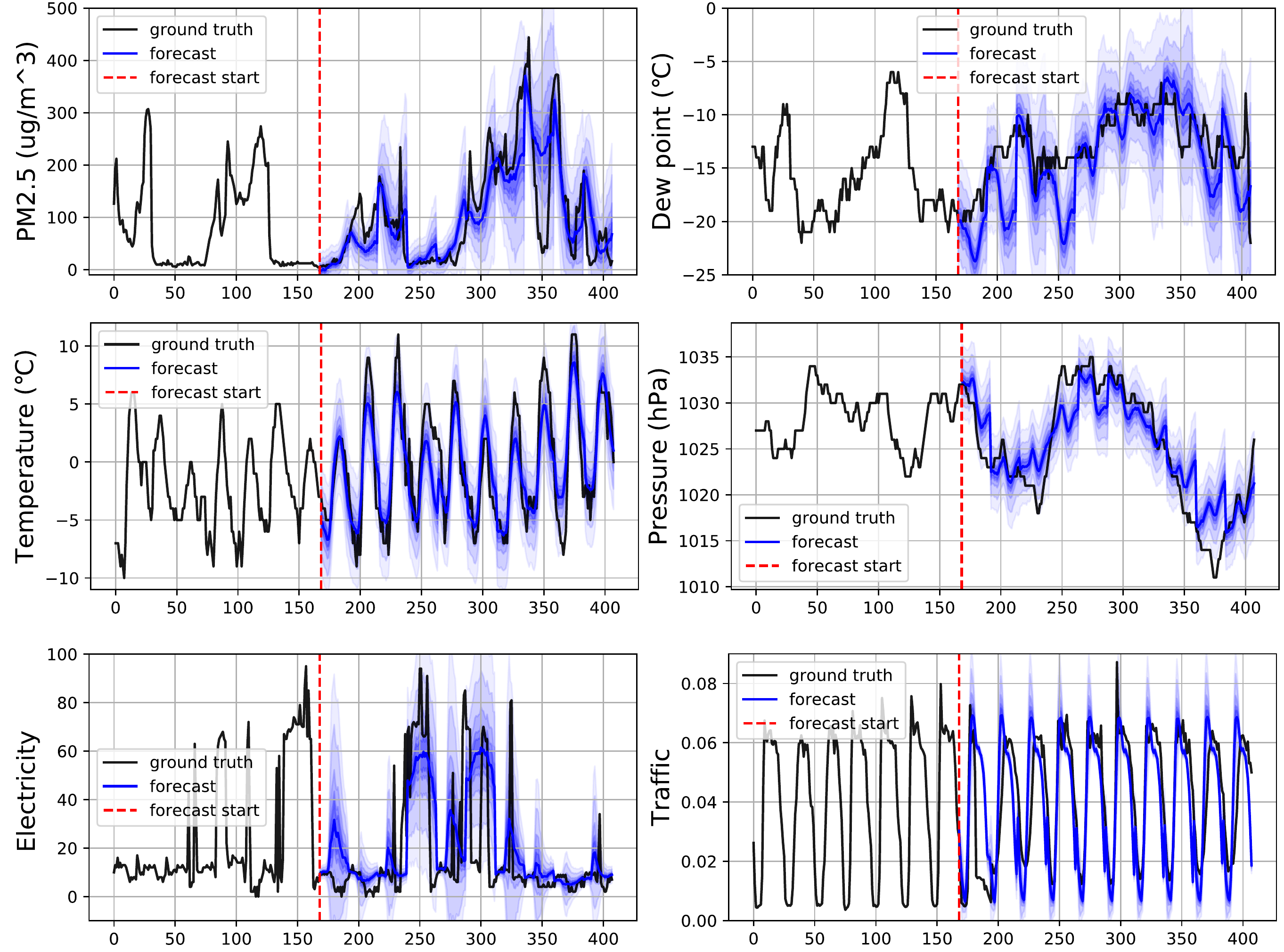}}
    \caption{Rolling-day probabilistic forecast by  VSMHN. Red vertical dashed lines show the start time of the forecast. Probability densities are plot by confidence intervals [20\%, 30\%,50\%, 80\%, 95\%] in transparency. The past series are not shown in full length. \textbf{The four variables from the first two rows i.e. PM2.5, Dew Point, Temperature, Pressure come from a multivariate dataset: Air Quality, and they are jointly modeled to obtain the shown results.}}
    \label{fig:visualization}
\end{figure}

\textbf{Forecast visualization.}
Figure \ref{fig:visualization} gives examples of probabilistic forecast results for each of the six variables in our dataset, we can see that VSMHN makes accurate and sharp probabilistic forecast overall. For each of the forecast horizon, the forecast uncertainty grows over time naturally. More importantly, the model spontaneously places higher uncertainty when forecast may not be accurate, e.g. $t=350$ for PM2.5 (top left), $t=220$ for Dew point (top right), and $t=170$ for Electricity (bottom left), which is a desired feature for downstream decision making processes.

\begin{table}[tb!]
\centering
\resizebox{\linewidth}{!}{
\begin{tabular}{lrrrrrr}
\hline
Method  & PM2.5   & Dewpoint    & Temperature     & Pressure   & Electricity    & Traffic     \\ \hline
DeepAR  &0.924&0.928&\tb{0.988}&0.679&0.879&0.907\\
DeepAR-E&0.822&0.946&0.910&0.689&0.581&0.889\\
MQRNN   &-0.12&0.062&-0.081&-0.022&0.738&0.436\\
DeepFactor&-0.235&0.062&-0.081&-0.022&0.738&0.436\\\hline
VAR&0.944&0.944&\tb{0.988}&0.965&0.923&0.562\\
ETS&0.969&0.933&\tb{0.988}&0.971&0.885&0.451\\
SARIMA&0.877&0.863&0.962&0.974&0.904&0.797\\\hline
VSMHN-TS&0.981&0.877&0.975&0.970&0.970&0.884\\
VSMHN&\tb{0.987}&\tb{0.971}&\tb{0.988}&\tb{0.995}&\tb{0.984}&\tb{0.918}\\\hline
\end{tabular}}
\caption{Quantification comparison by R2 of uncertainty calibration quality in the forecasting horizon.} 
\label{tab:calibration_performance}
\end{table}
\section{Conclusion}
We have presented a novel approach based on the deep conditional generative model to jointly learn from heterogeneous temporal sequences (i.e. time series and event sequences). To our best knowledge, this is one of the first works to provide a joint modeling framework for multi-horizon probabilistic forecasting. Empirically we have shown that our proposed model outperforms the state-of-the-art forecast models on six variables from three public datasets.
\section{Acknowledgements}
This research is supported by the National Key Research and Development Program of China (2020AAA0107600, 2018YFC0830400), Natural Science Foundation of China under Grant No. 61972250, U19B2035, and 72061127003, and the ECNU-SJTU joint grant from the Basic Research Project of Shanghai Science and Technology Commission (No. 19JC1410102). 
\section{Ethics Statement}
As far as we know, we are the first to define and address the problem of jointly modeling time-series and event sequences for multi-horizon time-series forecasting. For existing time-series forecasting applications, our research shows that reasonable modeling of related event data can effectively reduce the deviation of time-series forecasting, making downstream predictive analysis and decision-making more consistent with the goal of risk control. 

This research has many societal implications. For example, economists can study the impact of political events on economic indicators, formulating more predictive economic policies. Environmentalists can study the impact of events such as volcanic eruptions or industrial activities on the environment to understand climate change. Retailers can analyze the impact of promotions and other activities on sales to arrange stock and inventory.

\bibliography{main}

\end{document}